\documentclass[11pt,reqno]{amsart}
\usepackage{geometry}
\usepackage[dvipsnames]{xcolor}
\usepackage{graphicx}
\usepackage{amssymb}
\usepackage{epstopdf}
\title{Bayesian Brain: Computation with perception to recognize 3D objects}
\author{Kumar Sankar Ray}
\begin{document}
\maketitle

\vspace*{-.35in}
\begin{center}
\footnotesize
Department of Computer Engineering \& Application\\
GLA University, Mathura, UP, India\\
\texttt{kumarsankar.ray@gla.ac.in}
\end{center}

\begin{abstract}
We mimic the cognitive ability of Human perception, based on Bayesian hypothesis, to recognize view-based 3D objects. We consider approximate Bayesian (Empirical Bayesian) for perceptual inference for recognition. We essentially handle computation with perception.\\

\medskip
\noindent\textbf{Keywords:} Bayesian brain, perception, approximate bayesian, Empirical bayesian, view-based 3D objects.
\end{abstract}

\section{Introduction}

This paper deals with the concept of approximate Bayesian (Empirical Bayesian) for perceptual inference to recognize view based 3D objects which are projected as 2D images. Views are characterized by two attributes; view likelihood and view stability.

View likelihood is the probability of a certain view of a given 3D object observed as 2D image which is an aspect of the given 3D object from a particular viewing direction. View stability indicates what extent image changes if the viewing direction is slightly changed: it is the generic view. Both the phenomena are identical up\,to the prior probability of camera orientation and can be used to compare images.

A large number of psychological research is found on the role of aspect in object recognition. Koenderink and Van Doorn [1,2] first considered this idea of aspects of a 3D object Tarr and Kriegman [3] performed research on the definition of view in the context of human perception. This lends credence to the theory of using views as the basis for object recognition.

\section{The Bayesian brain hypothesis}

As proposed by [4] the bayesian brain hypothesis is a probability based concept to generate the cognitive ability of perception as a generative model. Based on the concept that the brain has a model of the world using sensory inputs [5] we try to mimic the cognitive ability of human perception for recognising 3D objects.

The brain is an inference machine that generates optimal hypothesis which can be viewed as posterior probability (belief about the world given the sensed data) and which can be represented by the product between likelihood of the sensed data of the given world and a prior (past experience about the world). Perception which is expressed as best hypothesis becomes the posterior probability which maps the perceived data into the belief about the world. Fig.~1 depicts the above mentioned concept of computation with perception. Human perception is full of uncertainty. Hence it is represented by probabilistic approach to Bayesian hypothesis.

Note that human brain does not execute any task using probability; but we try to mimic human computation with perception (outside the human brain) using Bayesian approach as depicted in Fig.~1.

\begin{figure}[h]
\fbox{\includegraphics{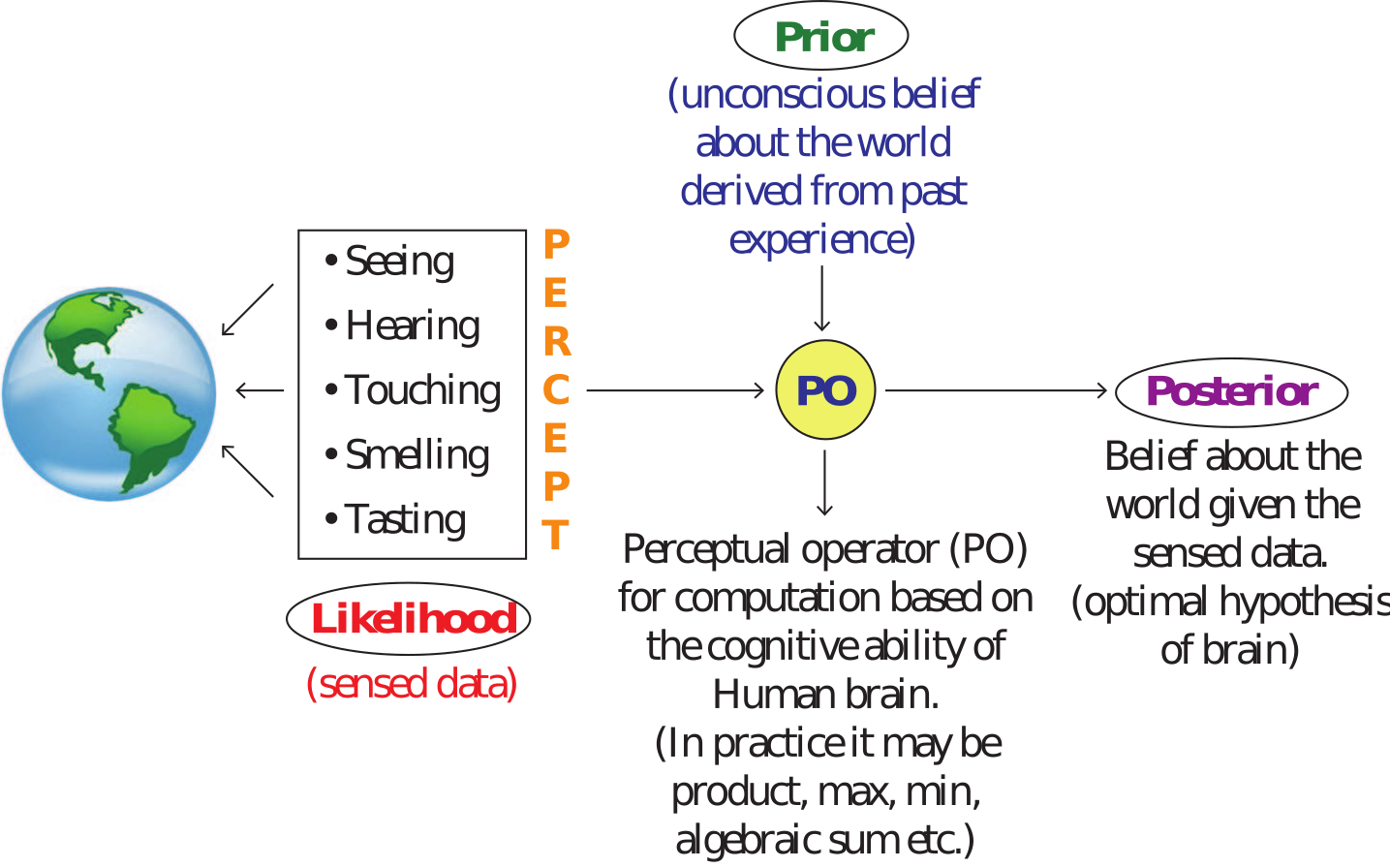}}
\caption{Computation with perception: a bayesian approach.}
\end{figure}

Computation with perception does not essentially deal with any specific numbers (numerical value). It is based on the resulting belief about the world obtained from the sensed data (percept). Hence in the design study we recognize an object based on the resulting belief about world of object represented by percept.

\section{Statement of the problem}

We compute the posterior probability of the world of 3D object based on the sensed data using equation (3).

We consider the 3D picture of a Kangaroo as shown in Fig.~2.

\begin{figure}[h]
\includegraphics[scale=.44]{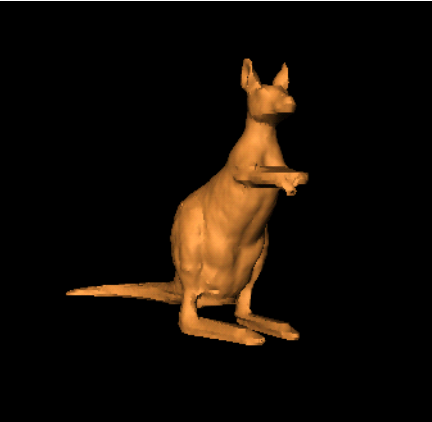}
\caption{Picture of kangaroo.}
\end{figure}

Different aspects of the Kangaroo are generated by taking projection of the object at regular interval (see Fig.~3). Kangaroo is placed inside the unit sphere.

\begin{figure}[h]
\fbox{\includegraphics[scale=.6]{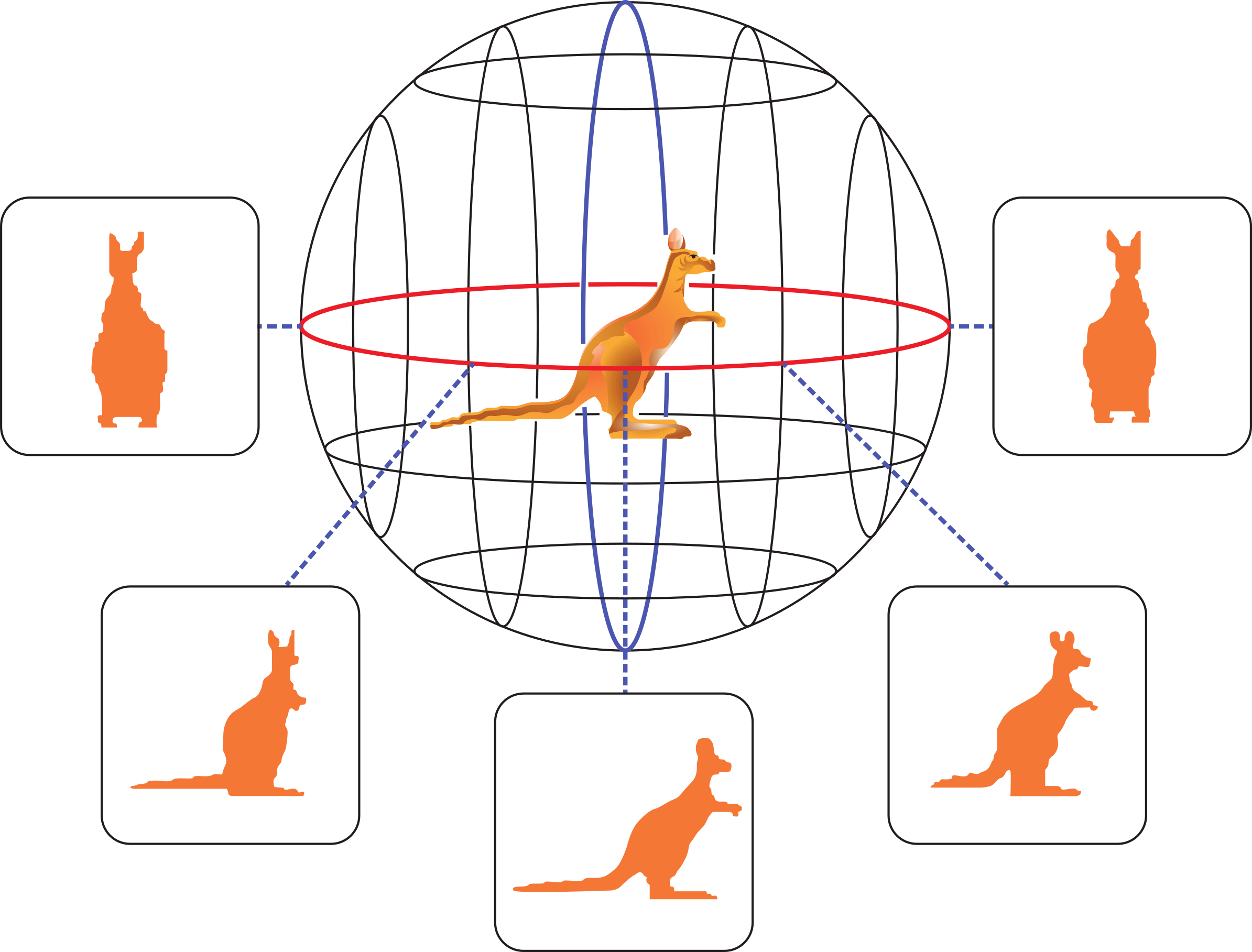}}
\caption{Different aspects of the kangaroo placed inside the unit sphere.}
\end{figure}

In Fig.~4 Kangaroo is shown on the left. On the right we show a range of views from the ground plane in five degree increment. Circled view represent prototypical views. Aspect boundaries are vertical lines in green. The posterior probability of 3D object, which is basically a Bayesian inference provides the cognitive ability of recognition in terms of computation with perception.

\begin{figure}[h]
\fbox{\includegraphics{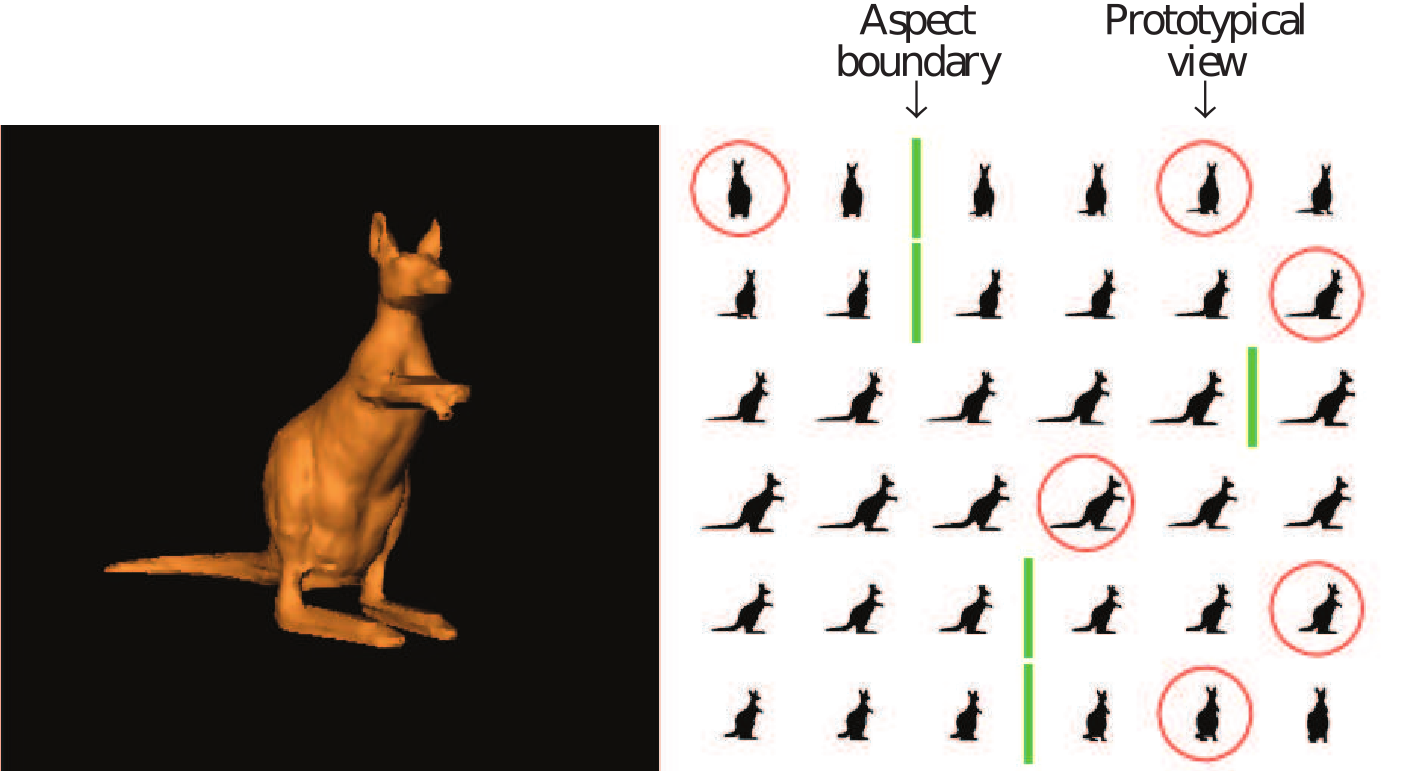}}
\caption{Different aspects of the kangaroo with prototypical views and aspect boundaries.}
\end{figure}

\section{Bayes rule}

Let us consider the Bayes rule,
\begin{equation}
\underbrace{P(B|A)}_{\footnotesize\begin{tabular}{@{}c@{}}Probability\\of B given that\\ A occurred\end{tabular}}\kern-10pt=\frac{P(A|B)\times\kern-10pt\overbrace{P(B)}^{\footnotesize\begin{tabular}{@{}c@{}}Probability\\of B\end{tabular}}}{\underbrace{P(A)}_{\text{Probability of A}}}
\end{equation}

The simplified form of Bayes rule is as follows,
\begin{equation}
P(B|A)\propto P(A|B)\times P(B)
\end{equation}
\begin{equation}
\Rightarrow\ P\ (\underbrace{\text{World $|$ Sensed data}}_{\footnotesize\begin{tabular}{@{}c@{}}\textcolor{Violet}{\textbf{Posterior}}\\(resulting beliefs\\about the world)\end{tabular}})\propto
P\ (\underbrace{\text{Sensed data $|$ world}}_{\footnotesize\begin{tabular}{@{}c@{}}\textcolor{red}{\textbf{Likelihood}}\\(world states generated\\ by the sensed data)\end{tabular}})\times
\kern-8pt
\underbrace{P\ (\text{world})}_{\footnotesize\begin{tabular}{@{}c@{}}\textcolor{Green}{\textbf{Prior}}\\(belief about the\\ world based on\\experience)\end{tabular}}
\end{equation}

According to Helmholtz [5], ``Perception is our bet hypothesis (Guess) as to what is in the world, given our current sensory evidence and our prior experience''.

As per Bayes rule, the prior distribution is fixed before any data are observed.

\subsection{Empirical Bayes}

In contrast to Bayesian method, Empirical Bayes produces statistical inference in which the prior probability distribution is estimated from the data.

Despite this difference in perspective, empirical Bayes may be viewed as an approximation of Bayesian approach as stated above.

\section{Deriving the Posterior using approximate Bayesian}

In approximate Bayesian which is basically empirical Bayesian we consider uncertainty of the parameter given the data.

\subsection{Representing a likelihood}

Considering binomial distribution we get probability of likelihood as
\begin{equation}
p(k|n,Q)={n\choose k}Q^{k}(1-Q)^{n-k}
\end{equation}
where $k$ indicates the number of successful recognition of 3D object from its aspects (which are 2D images captured by a digital camera) and $n$ indicates the total number of aspects of the 3D object.

\subsection{Representing a Prior}

For representation of Prior based on approximate Bayesian (Empirical Bayesian) we are guided by a design philosophy: we can recognize an object which we have seen earlier.

Thus the prior of approximate Bayesian (empirical Bayesian) is an unconscious belief about the world which we consider for design study.

Bayesian decision theory is a normative enterprise. It represents how we reason and take decision; not how we actually execute them.

Perceptual prior decides between rival hypotheses those are compatible with current proximal input. It matches the statistics of the world we consider for design study.

Perceptual prior is mutable. Mutability of prior is necessary for perception. If the prior is not properly tuned to the environment perception will be inaccurate. We may achieve perceptual illusion if low probability is assigned to prior.

Hence for present design study, we estimate initial prior probability using Beta distribution which provides some unconscious belief or vague knowledge obtained from past experience about the world. Once we achieve posterior probability as a product of likelihood and the prior as stated above we can update the initial prior by the posterior probability which is based on initial prior. Thus the prior is updated by process of mutation and becomes more informative about the world we handle.

\subsection{Computing Posterior probability}

Considering Bayes rule we compute the posterior probability $p(Q|n,k).$

We represent Posterior as follows,
\begin{equation}
\text{Posterior}=\frac{\text{Likelihood $\times$ Prior}}
                      {\text{Marginal Likelihood}}
\end{equation}

Using equation (4) and (5) we get as follows,
\begin{equation}
p(Q|n,k)=\frac{\left[{n\choose k}Q^{k}\cdot(1-Q)^{n-k}\right]\times\left[\frac{1}{B(a,b)}Q^{a-1}(1-Q)^{b-1}\right]}{p(k)}.
\end{equation}

$p(k)$ is constant because number of success full recognition of aspect is known.

Thus we arrive at equation (7) which is same as equation (2).
\begin{equation}
\text{Posterior $\propto$ Likelihood $\times$ Prior}
\end{equation}

\section{Experimental Result}

Using equation (4) we calculate the probability of likelihood data using aspects of the object as shown in Fig.~4.

We have collected total five hundred aspects of the 3D object of Kangaroo. We consider a batch of hundred aspects each time and calculate the probability of likelihood data. We consider each probability distribution of likelihood data as shown in each frame of Fig.~5. Next we compute the posterior probability as a product between each likelihood and prior [see equation (7)]. Thus we obtain total five posterior probability distributions. For simplicity of demonstration we map prior, likelihood and posterior on the same frame as shown in Fig.~5. In Fig.~5 we have shown only three frames out of five, i.e. the first frame, third frame and fifth frame. In the first frame we consider Beta $(4,4)$ for prior and derive the posterior as Beta $(84,24).$ We use the posterior as our prior for next experiment. In the fifth frame the posterior is Beta $(144,64).$ All the five posterior probability distributions have mean value which is closely clustered around 0.75 and thus the posterior probability is stable and no further iteration is necessary.

\begin{figure}[h]
\fbox{\includegraphics[scale=.5]{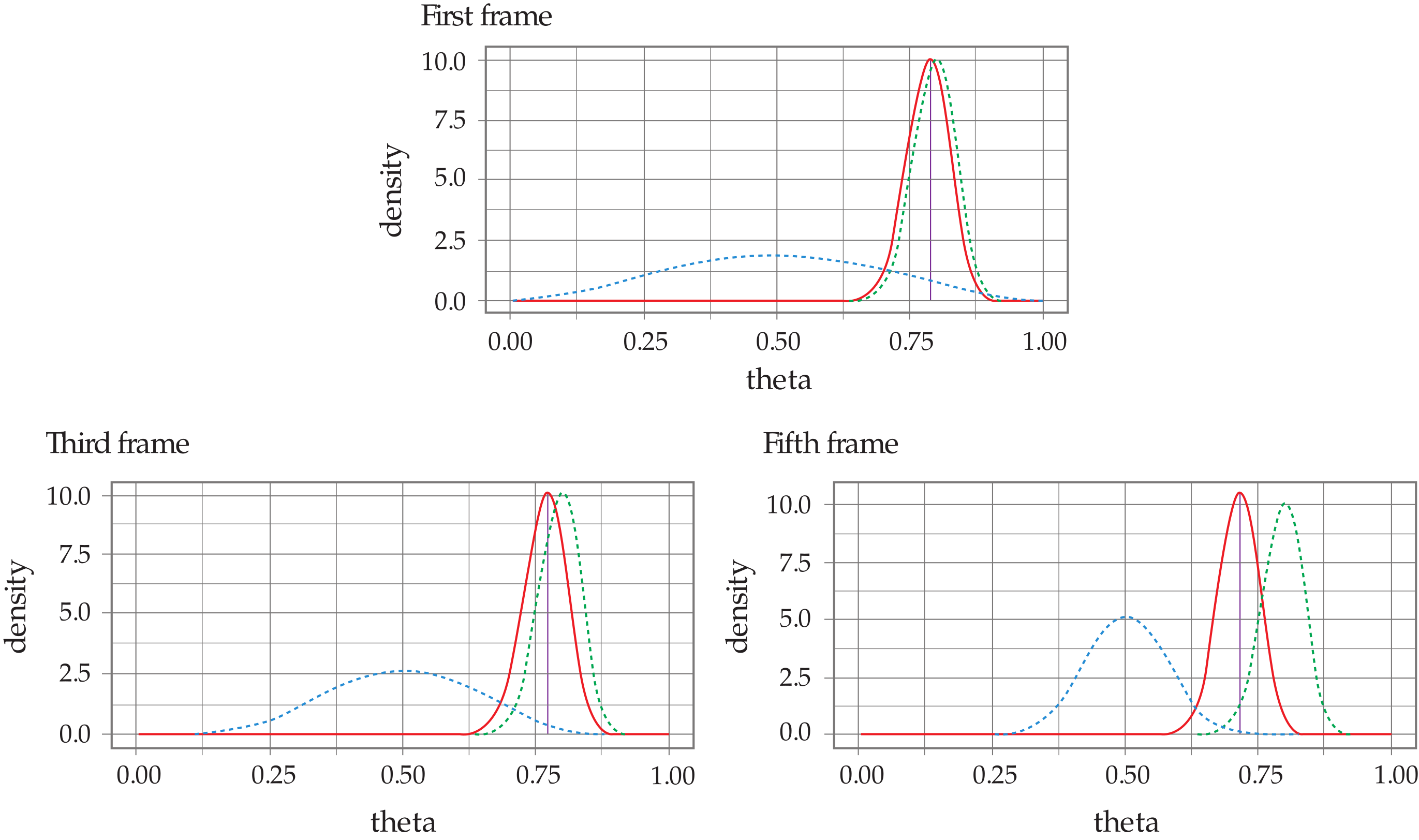}}
\caption{Recognition of 3D object from the posterior probability distribution of each frame is derived from the resulting belief about the world at each frame (\textcolor{red}{\textbf{-}} Posterior, \textcolor{Blue}{\textbf{-}} Prior, \textcolor{OliveGreen}{\textbf{-}} Scaled likelihood).}
\end{figure}

From five posterior probability distributions we obtain five inferences. Each inference provides information that the object is kangaroo. Thus we confirm recognition about 3D object (kangaroo) as shown in Fig.~4.

Each inference derived from posterior probability is basically a hypothesis which is a belief about the world [see equation (3)]. In this experiment recognition of 3D object is derived from the belief about the world as stated by the posterior probability of equation~(3). Thus the five hypotheses correctly recognize the object. In case we get less number of hypothesis for correct recognition we should consider at least 80\,\% of the total number of hypothesis of the experiment. For instance, in the present design study out of five posterior probability if four of them provide satisfactory hypothesis then computation with perception for recognition 3D object is acceptable. If it is less than four hypotheses then we should reformulate the design study.

\section{Conclusion}

We have successfully implemented computation with perception: a Bayesian approach. In stead of using Bayesian rule directly we consider empirical Bayes so that prior probability can be measured from the initial stage. In Bayes rule the product operation between likelihood and prior can be replaced by other operators like max, min, algebraic sum. Bayesian algorithm for high-dimensional estimation is a major challenge for computational neuroscientist.

\end{document}